\documentclass{bmvc2k}

\usepackage{amsmath}
\usepackage{amsfonts}
\usepackage{amssymb}
\usepackage{booktabs}
\usepackage{wrapfig}
\usepackage{comment}
\usepackage{mathrsfs}
\usepackage[capitalise, nameinlink]{cleveref} 

\usepackage{soul}
\usepackage[colorinlistoftodos,prependcaption,textsize=tiny]{todonotes}

\title{Uncertainty and Prediction Quality Estimation for Semantic Segmentation via Graph Neural Networks}

\newcommand\blfootnote[1]{%
  \begingroup
  \renewcommand\thefootnote{}\footnote{#1}%
  \addtocounter{footnote}{-1}%
  \endgroup
}

\addauthor{Edgar Heinert$^\ast$}{heinert@uni-wuppertal.de}{1}
\addauthor{Stephan Tilgner$^\ast$}{tilgner@uni-wuppertal.de}{2}
\addauthor{Timo Palm}{palm@math.uni-wuppertal.de}{1}
\addauthor{Matthias Rottmann}{rottmann@uni-wuppertal.de}{1}
\addinstitution{
 School of Mathematics and Natural Sciences\\
 University of Wuppertal\\
 Wuppertal, Germany
}

\addinstitution{
 School of Electrical, Information and Media Engineering\\
 University of Wuppertal\\
 Wuppertal, Germany
}

\runninghead{Heinert, Tilgner, Palm, Rottmann}{Graph-Based Meta Models}

\begin{document}

\maketitle

\blfootnote{$^\ast$ Equal contribution. This work has been supported by the German Federal Ministry of Education and Research within the junior research group project ``UnrEAL'', grant no.\ 01IS22069, and by the VolkswagenStiftung, ``DigiData'', grant no. 9C436.}

\begin{abstract}
When employing deep neural networks (DNNs) for semantic segmentation in safety-critical applications like automotive perception or medical imaging, it is important to estimate their performance at runtime, e.g.\ via uncertainty estimates or prediction quality estimates. Previous works mostly performed uncertainty estimation on pixel-level. In a line of research, a connected-component-wise (segment-wise) perspective was taken, approaching uncertainty estimation on an object-level by performing so-called meta classification and regression to estimate uncertainty and prediction quality, respectively. In those works, each predicted segment is considered individually to estimate its uncertainty or prediction quality. However, the neighboring segments may provide additional hints on whether a given predicted segment is of high quality, which we study in the present work. On the basis of uncertainty indicating metrics on segment-level, we use graph neural networks (GNNs) to model the relationship of a given segment's quality as a function of the given segment's metrics as well as those of its neighboring segments. We compare different GNN architectures and achieve a notable performance improvement.
\end{abstract}

\section{Introduction}
\label{sec:intro}
In recent years, DNNs for visual perception tasks have increasingly been used in safety-critical applications such as advanced driver assistance systems \cite{li2018real, jang2019semantic}, autonomous driving \cite{xiao2020multimodal, lee2023end, hoermann2018dynamic}, or medical diagnostics \cite{10.1007/978-3-319-24574-4_28, 7422783, kermi2019deep, https://doi.org/10.1002/mp.14609, 10466532}.
 
However, their decision-making process for predictions is mostly opaque, and the quality of a particular prediction is not quantified in standard semantic segmentation DNNs. Typically, the quality of such models is evaluated empirically by using collected data samples to test the model's predictive ability, resulting in a quantity averaged over the whole test set. When a semantic segmentation DNN is deployed for a given safety-critical task, it is of importance to have an indication of the performance of the DNN for the given data rather than an average performance over a greater test set \cite{hullermeier2021aleatoric}. Most of the early works in uncertainty estimation for semantic segmentation DNNs focused on a pixel-wise representation of the uncertainty estimates \cite{BMVC2017_57, NIPS2017_2650d608}
also rather focussing on a Bayesian point of view. It can be argued, however, that in practice a strict separation of aleatoric and epistemic uncertainty is not necessarily of importance \cite{hullermeier2021aleatoric}.

Inspired by the idea that an uncertainty estimate on an object level, rather than pixel level, would be desirable, a line of research pursues uncertainty estimates and prediction quality estimation on the level of class-wise connected components (segments) in the predicted semantic segmentation masks obtained by the DNN \cite{rottmann2019uncertainty, rottmann2020detection, chan2021entropy, chan2020controlled}.

\begin{figure*}
    \centering
    \includegraphics[trim={0 3cm 0 0}, width=.95\linewidth]{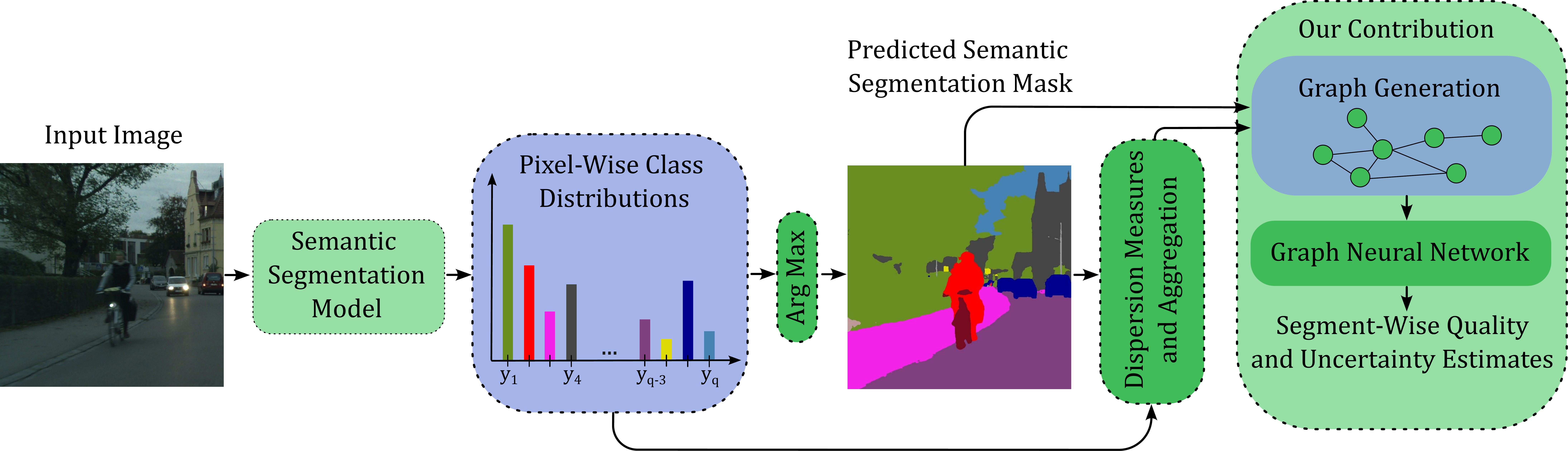}
    \caption{An illustration of our graph-based uncertainty estimation and prediction quality estimation approach.}
    \label{fig:pipeline}
    \vspace{-12pt}
\end{figure*}

These works focus on extracting features or hand-crafted metrics from the DNN's softmax output, therefrom performing a so-called meta classification and meta regression. Both tasks require a little hold out dataset to train on. In the given context, meta classification refers to discriminating true positive from false positive predictions and meta regression refers to the task of estimating the segment-wise intersection over union (IoU) of the given predicted segment with the ground truth. In both tasks, small-scale post-processing models perform predictions based on the hand-crafted metrics that are aggregated over the given segment under inspection. The works \cite{maag2020time,  fieback2023temporal} extend this post-processing framework by a temporal component. However, the spatial component has not been considered up to now. Neighboring segments may provide additional hints on whether a given predicted segment is of uncertain or of low quality, e.g.\ in street scenes it might be unlikely to observe a human in the sky.

To study the neighbor dependence in the tasks of meta classification and meta regression for semantic segmentation DNNs, we use different GNNs \cite{hamilton2017inductive, velivckovic2017graph} to model the neighbor relations of the predicted segments and their effect on the segment-wise IoU as well as the correctness of the given predicted segment. Those different GNN models include the utilization of basic aggregation functions \cite{hamilton2017inductive} and attention-mechanisms \cite{velivckovic2017graph}. The overall method is illustrated in \cref{fig:pipeline}.
We evaluate varying architectures of the different GNN types on the publicly available Cityscapes \cite{cordts2016cityscapes} dataset. The best GNN models outperform the baseline models, not considering any neighboring segments, by up to 1.78 percentage points in terms of AUROC in the classification task.

Our main contributions in this work are: \vspace{-0.5ex}
\begin{itemize}
    \item We incorporate information from neighboring predicted segments into meta classification and meta regression for semantic segmentation DNNs by modeling the spatial relationship between neighboring segments as edges in a graph. \vspace{-1ex}
    \item We perform a comprehensive study of the performance of several GNN architectures as well as models for semantic segmentation of street scenes. \vspace{-1ex}
    \item Our method outperforms baseline models that do not consider neighboring segments by up to 1.78 percentage points in the classification task. \vspace{-1ex}
\end{itemize}
We make our implementation publicly available under {\small\url{https://github.com/eheinert/Uncertainty-and-Prediction-Quality-Estimation-for-SemSeg-via-GNNs}.}

\section{Related Work}

Quality and uncertainty estimation for DNN's prediction are both closely related tasks, as a high quality prediction is often associated with low uncertainty. Accordingly, both tasks are addressed in the related work section, which is divided into two parts: pixel-wise and segment-wise approaches for semantic segmentation and related tasks.

\textbf{Pixel-Wise Uncertainty.}
One of the first uncertainty estimates at inference time in semantic segmentation models was Monte Carlo (MC) dropout proposed in \cite{BMVC2017_57} as an extension of \cite{gal2016dropout}. The authors employ Monte Carlo sampling, implemented through the use of dropout \cite{JMLR:v15:srivastava14a} during inference, to estimate uncertainty for each pixel of the model's prediction. Dropout for uncertainty estimation for DNNs was initially proposed in \cite{gal2016dropout}.
A variety of approaches \cite{NIPS2017_2650d608, devries2018leveraginguncertaintyestimatespredicting, Sander_2019} utilize an uncertainty measure in the prediction of semantic segmentation models through MC dropout. However, this approach has the disadvantage of requiring multiple inferences on an image to estimate the uncertainty. In real-time applications such as autonomous driving, the computational burden associated with the utilization of MC dropout as an uncertainty estimation method may be considerable.

In \cite{ KARIMI2019186}, the authors propose to use a five ensemble for clinical target volume segmentation of the prostate in ultrasound images, providing a predictive pixel-wise quality estimation. Therein, the pairwise Dice Similarity Coefficient \cite{ https://doi.org/10.2307/1932409} of the ensemble predictions is used as a quality estimate. In \cite{9607788}, knowledge distillation methods \cite{hinton2015distillingknowledgeneuralnetwork} transfer uncertainty estimates from a teacher ensemble segmentation to a single student model, eliminating the necessity for multiple inferences. This has the effect of reducing the computational burden associated with estimating prediction uncertainty during inference.

A semantic segmentation approach that leverages hyperbolic embeddings instead of the conventional Euclidean ones is proposed in \cite{ atigh2022hyperbolic}, providing a pixel-wise quality estimate.
All of these methods ultimately yield pixel-wise or (via averaging) image-wise uncertainty measures or quality estimates.

\textbf{Segment-Wise Uncertainty.}
Initially, \cite{rottmann2020prediction} proposed a methodology for estimating the quality and uncertainty of each segment of a segmentation model's prediction, as opposed to the pixel or image-level. Estimates at the segment-level are more conducive to integration into a downstream automated decision-making process, such as that employed in autonomous driving. The core concept of this methodology is the aggregation of per-pixel dispersion measures, such as entropy and probability differences, pertaining to the predicted class distribution, in conjunction with additional geometric information regarding the segment, including the number of pixels, as features. Based on segment-wise features carrying uncertainty, the segment-wise IoU with the ground truth is estimated (termed meta regression) as well as the probability of a correct prediction (termed meta classification). The entire framework is termed MetaSeg.
A variety of methods make use of MetaSeg for the purpose of detecting or reducing false positive and false negative samples \cite{rottmann2020detection, chan2020controlled, maag2022false}, identifying out-of-distribution objects \cite{ oberdiek2020detection, bruggemann2020detecting, chan2021entropy}, implementing active learning strategies \cite{colling2020metabox+}, anomaly detection \cite{chan2022detecting, uhlemeyer2022towards}, detecting label errors in annotated datasets \cite{rottmann2023automated} or serving for segment-wise analysis purposes \cite{heinert2024reducing,rosenzweig2021validation}.

MetaSeg is employed in \cite{rottmann2019uncertainty} on nested image crops of an image, yielding a batch of pixel-wise class distributions, which are combined before applying MetaSeg. Using multiple crops enhances the performance of the segmentation model and of the meta models - a term we use to refer to both meta classification as well as regression models.
In \cite{maag2020time}, an extension of MetaSeg incorporates temporal aspects, integrating the segment-wise uncertainty derived from preceding frames (in the context of video data) into the current frame's segment-wise estimation of uncertainty and quality.
Furthermore, temporal MetaSeg is employed within the domain of instance segmentation \cite{maag2021improving,maag2021false}. Also extensions of MetaSeg to 2D objects detection \cite{schubert2021metadetect,riedlinger2023gradient}, 3D object detection \cite{riedlinger2023lmd} and 3D Lidar point cloud segmentation \cite{colling2021false,colling2023prediction} are available.
However, none of the approaches to uncertainty estimation at the segment-level make use of information from neighboring segments.

\section{Neighbor-Aware Meta Classification and Regression}

\begin{wrapfigure}{r}{0.7\textwidth}
    \vspace{-8pt}
    \centering
    \includegraphics[trim=0 0 0 0,clip,width=0.9\linewidth]{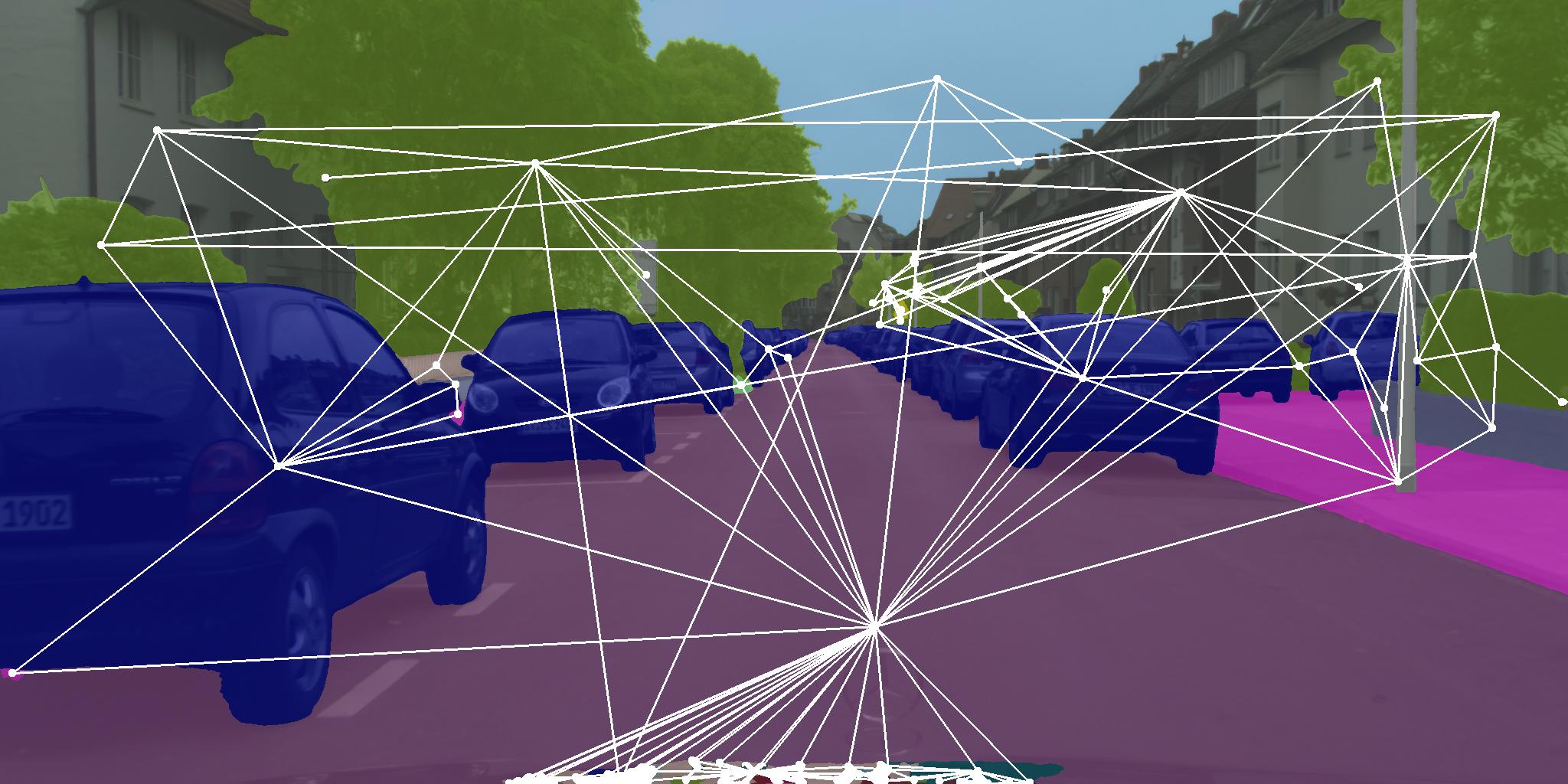}
    \caption{Cityscapes image overlayed with a graph generated from segment barycenters. Each segment is connected to adjacent ones}
    \label{fig:graph_on_street}
    \vspace{-6pt}
\end{wrapfigure}

Our approach extends a semantic segmentation post-processing method by Rottmann et al. \cite{rottmann2020prediction}, which estimates the quality of each predicted segment using metrics, or in other words constructed features that carry uncertainty information, derived from the softmax output. However, \cite{rottmann2020prediction} treats the derived metrics of each segment independently and does not utilize information about a segment's neighborhood. To incorporate neighborhood information of predicted segments, we introduce a data structure that captures this information and apply meta-estimators capable of leveraging it. The natural choice for such a data structure is an undirected graph representing the information of each frame. For the estimators, we employ inductive GNNs, such as GraphSAGE and Graph Attention Networks (GAT), which are suitable for graphs of arbitrary shapes.

\textbf{Graph and Feature Construction.}
In this paragraph, except for the graph construction, we mainly follow \cite{rottmann2020prediction}. For an image $x \in [0,1]^{h \times w \times d}$ and one of its pixels $z$, the ground truth semantic segmentation with $q+1$ classes is $y \in \mathcal{C}= \{y_1,\ldots,y_{q+1}\}$ with class $y_{q+1}$ being unlabeled. The pixel-wise raw softmax output of a semantic segmentation network for $z$, $f_z(y\,|\,x,w)$ is a probability distribution over the first $q$ classes $\mathcal{C}'$ and the networks pixel-wise prediction $\hat{y}_z\in \mathcal{C}'$ is given by $\hat{y}_z = {\text{arg\,max}}_{y \in \mathcal{C}'}\,f_z(y\,|\,x,w).$ 

Let $\mathcal{\widehat{K}}_x=\{\hat{k}_1, \dots, \hat{k}_M\}$ be the set of predicted segments and $\mathcal{K}_x$ the set of ground truth segments of a given image. Herein, a segment refers to a connected component of a given class in a segmentation mask where all eight surrounding pixels are considered to be adjacent. We then define the image-wise graph $\mathbf{G}_x=(\mathcal{\widehat{K}}_x, \mathcal{\widehat{E}}_x)$ with $\widehat{e}_{l,m}:=\{\hat{k}_l,\hat{k}_m\} \in \mathcal{\widehat{E}}_x$, iff a pixel in $\hat{k}_l$ neighbors a pixel in $\hat{k}_m$.
An illustration of such a graph can be found in \cref{fig:graph_on_street}. As before, for each predicted segment $\hat{k}$ in $\mathcal{K}_x$, a number of scalar metrics $h \in \mathbb{R}^p$ are derived, which are the node-wise input vectors of our meta estimators. The metrics are based on geometric properties of the respective predicted segments like the set of inner pixels $\hat{k}_{in}$, the boundary $\hat{k}_{bd}=\hat{k} \setminus \hat{k}_{in}$, as well as dispersion metrics like the pixel-wise entropy and difference in two highest probabilities,
\begin{align} 
    E_z(x, w) &= -\frac{1}{\log(q)} \sum_{y \in \mathcal{C}'} f_z(y\,|\,x, w) \log f_z(y\,|\,x, w), \\
D_z(x, w) &= 1 - f_z\big(\hat{y}_z(x, w)\,|\,x, w\big) \hspace{+16pt} + \hspace{-0pt} \max_{y \in \mathcal{C}' \setminus \big\{\hat{y}_z(x, w)\big\}} \hspace{-14pt} f_z(y\,|\,x, w). \vspace{-3ex}
\end{align}
Further metrics can be found in \cite{rottmann2020prediction,chan2020controlled} and our implementation as we only present an excerpt: 
\begin{itemize}
\item the number of all segments $|\hat{\mathcal{K}}_x|$ and the specific segment's barycenter. $(\bar{h}_{\hat{k}},\bar{w}_{\hat{k}})$ \vspace{-1ex} 
    \item the number of all, inner, and boundary pixels of a segment and their relations $S=|\hat{k}|$, $S_{in}=|\hat{k}_{in}|$, $S_{bd}=|\hat{k}_{bd}|$, $S_{rel}=S/S_{bd}$ etc. \vspace{-1ex}
    \item average entropies and probability distances $\bar{E}$, $\bar{E}_{in}$, $\bar{E}_{bd}$, $\bar{D}$, $\bar{D}_{in}$, $\bar{D}_{bd}$ defined as
    \vspace{-4pt}
    \begin{align}
        \bar{E}_\# = \frac{1}{S_\#} \sum_{z \in \hat{k}_\#} E_z(x), \; \bar{D}_\# = \frac{1}{S_\#} \sum_{z \in \hat{k}_\#} D_z(x) \quad \text{for } \# \in \{\_, \text{in}, \text{bd}\}
    \end{align}
    \vspace{-14pt}
    
    and analogous variances. \vspace{-1ex}
    \item the class-wise average softmax probabilites
    $\bar{p}_y = \frac{1}{S} \sum_{z \in \hat{k}} f_z(y\,|\,x,w) \text{ for }y \in \mathcal{C}'$. \vspace{-1ex}
\end{itemize}
For $\mathcal{K}_x|_{\hat{k}}$, the set of ground truth segments that intersect with $\hat{k}$ and have the same ground truth class as the predicted class for $\hat{k}$, $K' = \bigcup_{k' \in \mathcal{K}_x|_{\hat{k}}} k'$ and $Q = \left\{ q \in \hat{\mathcal{K}}_x \setminus\{\hat{k}\} : q \cap K' \neq \emptyset \right\}$ we finally define the adjusted IoU \cite{rottmann2020prediction}\vspace{-6pt}
\begin{equation}
    \text{IoU}_{\text{adj}}(\hat{k}) = \frac{|\hat{k} \cap K'|}{|\hat{k} \cup (K' \setminus Q)|}\vspace{-6pt}
\end{equation}
as the target metric. 

\textbf{Meta Classification and Regression.}
Meta regression and classification are both segment-wise tasks. For each predicted segment, we perfom two learning tasks. \emph{Meta Regression} refers to the task of predicting $\text{IoU}_{\text{adj}}(\hat{k})$ for a given segment $\hat{k}$ based on its metrics / constructed features introduced in the previous subsection. \emph{Meta Classification} refers in analogy to meta regression to a corresponding classification task where the  $\text{IoU}_{\text{adj}}(\hat{k})$ is replaced by the binarized target quantity\vspace{-8pt}
\begin{equation}
    \text{IoU}_0(\hat{k}) =
\begin{cases}
  0 & \text{if } \quad \text{IoU}_\text{adj}(\hat{k}) = 0, \\
  1 & \text{if } \quad \text{IoU}_\text{adj}(\hat{k}) > 0 \, .
\end{cases}
\end{equation}\vspace{-8pt}

While meta regression aims at estimating prediction quality of localization quality for a predicted segment, meta classification aims at estimating the uncertainty about the correctness of the prediction $\hat{k}$.

\textbf{Graph Neural Networks.}
In the following we use the metrics from above as covariables to estimate certain quantities, giving rise to uncertainty as well as prediction / localization quality. To also incorporate the features of neighboring predicted segments, we utilize GNNs. In the following we introduce the types of GNN layers used in our method.

\emph{GraphSAGE} \cite{hamilton2017inductive} was designed for inductive learning, which means that the model is generally applicable to undirected finite graphs of arbitrary shape and, therefore, aligns well with the objective of meta classification and meta regression. The fundamental concept of a GraphSAGE layer is the aggregation of node features from a uniformly sampled subset of all one-hop neighbors for each node, which are then concatenated with the node's own features. In this context, the term \emph{$k$-hop neighbor} is defined as a node that can be reached via the shortest path of length $k$ from the node under consideration. The input-output characteristic of a layer for each node $\hat{k} \in \mathcal{\hat{K}}_x$ is described by (cf.\ \cite{hamilton2017inductive})
\begin{subequations}
\begin{align}
    h_{\tilde{\mathcal{N}}(\hat{k})} &= \text{AGGREGATE}\Big(\{h_{\hat{v}} \: | \: \hat{v} \in \tilde{\mathcal{N}}(\hat{k})\}\Big), \\
    \tilde{h}_{\hat{k}} &= \sigma\Big(W \cdot \text{CONCAT}\big(h_{\hat{k}}, h_{\tilde{\mathcal{N}}(\hat{k})}\big) \Big),
\end{align}
\end{subequations}
where $h_i \in \mathbb{R}^p$ is the feature vector of node $i$, $W$ a learnable weight matrix, $\sigma$ an activation function, and the function $\tilde{\mathcal{N}}(\hat{k})$ gives a uniformly sampled subset of all one-hop neighborhood node indices of node $\hat{k}$. As the set of one-hop neighbors feature vectors is unordered, the authors propose the use of aggregation functions that are invariant to permutations of their inputs. Examples of such functions include the mean and the maximum.

\emph{Graph Attention Networks (GATs)} \cite{ velivckovic2017graph} are analogous to GraphSAGE in that they can be utilized for inductive learning and leverage the local neighborhood structure. However, the fundamental concept is to employ an attention mechanism \cite{bahdanau2014neural} for the aggregation of the one-hop neighbor's features. A pairwise attention weight $\alpha_{\hat{k}\hat{v}}$  between a node $\hat{k} \in \mathcal{\hat{K}}_x$  and its one-hop neighbors $\hat{v} \in \mathcal{N}(\hat{k}) \cup \{\hat{k}\}$ (including $\hat{k}$) is employed to form the output (cf.\ \cite{ velivckovic2017graph})
\begin{align}
\label{eq_gat}
    \tilde{h}_{\hat{k}} = \sum_{\hat{v} \in \mathcal{N}(\hat{k}) \cup \{\hat{k}\}} \alpha_{\hat{k}\hat{v}} W h_{\hat{v}}
\end{align}

of the GAT layer for each node. This is extended to a multi-head attention mechanism, where \cref{eq_gat} is applied K-fold in parallel with different weight matrices $\{W^l\}$ and attention weights $\{\alpha_{\hat{k}\hat{v}}^l\}$. The resulting set $\{\tilde{h}_{\hat{k}}^l\}$ is then averaged to produce the final output $\tilde{h}_{\hat{k}}$.

\section{Numerical Experiments}

In this section, we investigate the performance of our method and present an analysis of the impact of the choice of hyperparameters, such as depth, width, layer types, and learning rate, of graph-based meta classification and meta regression models. In all experiments, a fixed 80/20 split of the validation set of the publicly available Cityscapes dataset \cite{cordts2016cityscapes} was employed for training and evaluation of the meta models. The validation set comprises 500 pixel-wise labeled street scene images with a resolution of 2048 $\times$ 1024 pixels. Moreover, we utilized the DeepLabv3+ \cite{ chen2018encoder} network for semantic segmentation with two backbones: WideResNet38 \cite{wu2019wider} and ResNet18 \cite{he2016deep}. For the sake of brevity, they will be referred to as DV3+WRN38 and DV3+RN18, respectively. We use trained checkpoints from \cite{zhu2019improving} for DV3+WRN38, while we trained DV3+RN18 with only 50 annotated images to obtain a model with low mIoU. The model DV3+WRN38 demonstrates excellent semantic segmenation performance with a mIoU of 83.45\% on the Cityscapes validation set. DV3+WRN38 reaches a comparatively low mIoU of 46.29\% on the same validation data. Our graph-based meta model employs a combination of GAT, GraphSAGE, and linear layers (fully connected layers). Specifically, a Bayesian optimization method from the SMAC3 package \cite{JMLR:v23:21-0888} was employed to identify optimal model architectures by using the validation data of Citycapes. 

The remainder of this section is organized in the following manner: First, the results of the graph-based meta classification and graph-based meta regression experiments are shown in a baseline comparison. Thereafter, the process of selecting the GNN architectures for the graph-based meta estimators is described in detail, and an analysis of the impact of the choice of hyperparameters is provided. Finally, a comprehensive examination of the performance of the meta models with a particular focus on class-wise evaluation is presented. 

\textbf{Evaluation Procedure.} 
All model architectures (cf.\ \cref{tab:architectures_all}) identified in the selection process were trained five times to account for the randomness inherent in the learnable parameter initialization process and the resulting randomness in the optimization trajectory in parameter space.  Moreover, the Mean Square Error (MSE) loss function and the Adam optimizer \cite{ kingma2014adam} were employed during the training phase. The learning rates (cf.\ \cref{tab:architectures_all}) identified by the Bayesian optimization process for selecting the model architectures were utilized for training. The meta models were trained for 200 epochs and the performance metrics of the best model instances within each run are considered for evaluation. In addition, the performance of our proposed graph-based meta models is compared to a baseline comprising a set of models that do not incorporate information from neighboring segments. The baseline models are: logistic and linear regression as employed in the original MetaSeg framework \cite {rottmann2020prediction}, a gradient boosting model, GNNs without neighbors (in essence fully connected neural networks), and the two best-performing GNNs, which are utilized with graphs without edges. All meta models have been implemented in Python by using either Pytorch Geometric (PyG) \cite{fey2019fast} or Scikit-learn \cite{pedregosa2011scikit}.

\begin{table}
\centering
\scalebox{0.8}{
\begin{tabular}{llllll}
\toprule
\textbf{Model} & \textbf{LR} & \textbf{1. Layer} & \textbf{2. Layer} & \textbf{3. Layer} & \textbf{4. Layer} \\
\midrule
LLLL\textsubscript{C} & 0.021 & Linear (77) & Linear (57) & Linear (144) & Linear (1)\\
\midrule
GSGL\textsubscript{C} & 0.015 & GAT (91) & SAGE-mean (93) & GAT (154) & Linear (1) \\
LLSG\textsubscript{C} & 0.007 & Linear (293) & Linear (39) & SAGE-mean (179) & GAT (1)\\
LS\textsubscript{C} & 0.002 & Linear (242) & SAGE-mean (1) & - & - \\
LSS\textsubscript{C} & 0.009 & Linear (373) & SAGE-mean (269) & SAGE-mean (1) & - \\
SSS\textsubscript{C} & 0.002 & SAGE-mean (371) & SAGE-mean (283) & SAGE-mean (1) & - \\
LLSS\textsubscript{C} & 0.013 & Linear (304) & Linear (57)  & SAGE-mean (161) & SAGE-mean (1) \\
LLSL\textsubscript{C} & 0.015 & Linear (117) & Linear (22) & SAGE-mean (154) & Linear (1)\\
\midrule
\midrule
LL\textsubscript{R} & 0.001 & Linear (317) & Linear (1)& - & - \\
\midrule
LS\textsubscript{R} & 0.001 & Linear (314) & SAGE-mean (1) & - & - \\
LLS\textsubscript{R} & 0.006 & Linear (145) & Linear (103) & SAGE-mean (1) & - \\
LSS\textsubscript{R} & 0.009 & Linear (142) & SAGE-mean (137) & SAGE-mean (1) & - \\
LLSL\textsubscript{R} & 0.003 & Linear (337) & Linear (292) & SAGE-mean (99) & Linear (1) \\
\bottomrule
\end{tabular}}
\caption{Selected meta classification (top) and meta regression (bottom) architectures. Learning rate (LR) and the layer configuration of each selected model architecture are presented. The number of neurons in each layer is indicated in parentheses.}
\label{tab:architectures_all}
\vspace{-12pt}
\end{table}

\textbf{Graph-Based Meta Classification. }
In order to evaluate the meta classification models that estimate the segment-wise uncertainty of the segmentation models' predictions, the area under the receiver operating characteristic curve (AUROC) \cite{bradley1997use} and the F1-score \cite{ van1979information, christen2023review} are employed as performance metrics. The results are presented in \cref{tab:model_classification_performance} where in each row the means and the standard deviations of the performance metrics of the five model instances are given. Furthermore, the results of the five baseline models are presented in the top section of the table, while the GNNs' utilising neighboring segments' information are at the bottom.

\begin{table}
\centering
\scalebox{0.8}{
\begin{tabular}{lllll}
& \multicolumn{2}{c}{DV3+WRN38} & \multicolumn{2}{c}{DV3+RN18}\\
\toprule
\textbf{Model} & \textbf{AUROC} (\%) & \textbf{F1-Score} (\%) & \textbf{AUROC} (\%) & \textbf{F1-Score} (\%) \\
\midrule
logistic regression & 95.06 $\pm$ 0.00 & 82.09 $\pm$ 0.04 & 83.22 $\pm$ 0.00 & 65.11 $\pm$ 0.00\\
gradient boosting & 95.95 $\pm$ 0.00 & 83.54 $\pm$ 0.00 & 84.46 $\pm$ 0.00 & 66.06 $\pm$ 0.00\\
LLLL\textsubscript{C} & 95.87 $\pm$ 0.02 & 83.46 $\pm$ 0.08 & 84.59 $\pm$ 0.04& 66.57 $\pm$ 0.32 \\
LS\textsubscript{C,no edges} & 95.95 $\pm$ 0.02 & 83.48 $\pm$ 0.12 & 84.69 $\pm$ 0.06 & 66.73 $\pm$ 0.40\\
LSS\textsubscript{C,no edges} & 95.89 $\pm$ 0.01 & 83.33 $\pm$ 0.08 & 84.65 $\pm$ 0.04 & 66.08 $\pm$ 0.26\\
LLSS\textsubscript{C,no edges} & 95.93 $\pm$ 0.01 & 83.51 $\pm$ 0.18 & 84.65 $\pm$ 0.09 & 66.32 $\pm$ 0.54 \\
\midrule
GSGL\textsubscript{C} & \textit{87.95 $\pm$ 12.32} & \textit{64.06 $\pm$ 32.05} & \textit{80.79 $\pm$ 0.10} & \textit{59.43 $\pm$ 1.12}\\
LLSG\textsubscript{C} & 95.20 $\pm$ 0.58 & 82.48 $\pm$ 1.25 & 83.11 $\pm$ 0.58 & 65.31 $\pm$ 0.88\\
LS\textsubscript{C} & 96.10 $\pm$ 0.02 & 83.72 $\pm$ 0.11 & \textbf{85.00 $\pm$ 0.06} & \textbf{67.14 $\pm$ 0.26}\\
LSS\textsubscript{C} & \textbf{96.11 $\pm$ 0.02} & \textbf{83.81 $\pm$ 0.21} & 84.94 $\pm$ 0.04 & 66.69 $\pm$ 0.30\\
SSS\textsubscript{C} & 96.04 $\pm$ 0.02 & 83.65 $\pm$ 0.30 &84.73 $\pm$ 0.42& 66.22 $\pm$ 0.56 \\
LLSS\textsubscript{C} & 96.10 $\pm$ 0.03 & 83.77  $\pm$ 0.10 & 84.83 $\pm$ 0.12 & 66.77 $\pm$ 0.51\\
LLSL\textsubscript{C} & 96.05 $\pm$ 0.02 & 83.69 $\pm$ 0.06 & 84.84 $\pm$ 0.06 & 66.35 $\pm$ 0.29\\
\bottomrule
\end{tabular}}
\caption{Evaluation results of the meta classification models for WideResNet38 (column 2+3) and ResNet18 (column 4+5) as respective backend of the DeepLabV3+ segmentation model. The results are averages over five runs and the numbers are in percent.}
\label{tab:model_classification_performance}
\vspace{-12pt}
\end{table}

The best-performing model for the DV3+WRN38 case is the LSS\textsubscript{C} GNN with an average AUROC of 96.11\% and a mean F1-score of 83.81\%. The LSS\textsubscript{C} model is composed of a linear layer followed by two GraphSAGE layers with mean aggregation functions (cf.\ \cref{tab:architectures_all}). All baseline models are outperformed by LSS\textsubscript{C}, with the largest difference being 1.05 percent points relative to the logistic regression model and the smallest difference being 0.16 percent points relative to the gradient boosting model. It should be noted that five out of the seven GNN-based models demonstrate superior performance compared to the strongest baseline models, namely the gradient boosting and LS\textsubscript{C, no edges}. Furthermore, the standard deviations are notably small, with the exception of the GNNs that contain GAT layers, specifically GSGL\textsubscript{C} and LLSG\textsubscript{C}. Indeed, the GSGL\textsubscript{C} model, comprising two GAT layers, exhibits the poorest performance among all models, while LLSG\textsubscript{C}, which contains a single GAT layer, outperforms only the logistic regression baseline.

A comparable image can be derived from the evaluation results for the weaker DV3+RN18 model. Five of the seven GNNs demonstrate superior performance compared to all baseline models. The LS\textsubscript{C} GNN exhibits the highest performance, achieving a mean AUROC of 85\% and an average F1-score of 67.14\%. The difference between this result and that of the weakest baseline, namely logistic regression, is 1.78 percentage points. As was the case with the DV3+WRN38 result, the GNNs containing GAT layers perform the worst.  

\textbf{Graph-Based Meta Regression. }
In the evaluation of the graph-based meta regression models that estimate the segment-wise prediction quality of a segmentation model, the coefficient of determination (R\textsuperscript{2}-score cf.\ \cite{chicco2021coefficient}) is employed as a performance metric. As with the meta classification case, all GNN architectures for graph-based meta regression (cf.\ \cref{tab:architectures_all}) that were identified in the selection process, were trained and evaluated five times and the means and the standard deviations of the R\textsuperscript{2}-scores are reported. The results for the DV3+WRN38 and DV3+RN18 cases are presented in \cref{tab:model_r2_performance_regression}.

The best-performing meta regression model in the DV3+WRN38 case with an average R\textsuperscript{2}-score of 85.71\% is the baseline model LL\textsubscript{R} that is composed of two linear layers. The LS\textsubscript{R} model, which is the best-performing GNN that considers information from 
\begin{wraptable}{r}{0.57\textwidth}
\vspace{-4pt}
\centering
\scalebox{0.8}{
\begin{tabular}{lll}
& DV3+WRN38 & DV3+RN18\\
\toprule 
\textbf{Model} & \textbf{R\textsuperscript{2}-Score} (\%) & \textbf{R\textsuperscript{2}-Score} (\%) \\
\midrule
Linear Regression & \textit{83.45} & \textit{74.59} \\
Gradient Boosting & 85.05 $\pm$ 0.00 & 77.41 $\pm$ 0.00 \\
LL\textsubscript{R} & \textbf{85.71 $\pm$ 0.04} & 78.27 $\pm$ 0.07 \\
LS\textsubscript{R, no edges} & 85.67 $\pm$ 0.03 & 78.25 $\pm$ 0.07 \\
LLSL\textsubscript{R, no edges} & 85.32 $\pm$ 0.03 & 77.92 $\pm$ 0.14 \\
\midrule
LS\textsubscript{R} & 85.58 $\pm$ 0.04 & \textbf{78.29 $\pm$ 0.06} \\
LLS\textsubscript{R} & 85.39 $\pm$ 0.04 & 77.97 $\pm$ 0.10 \\
LSS\textsubscript{R} & 85.33 $\pm$ 0.07 & 53.05 $\pm$ 43.07 \\
LLSL\textsubscript{R} & 85.47 $\pm$ 0.09 & 78.20 $\pm$ 0.10 \\
\bottomrule
\end{tabular}}
\caption{Evaluation results of the meta regression models for WideResNet38 (column 2) and ResNet18 (column 3) as respective backend of the DeepLabV3+ segmentation model. The results are averages over five runs and displayed in percent.}
\label{tab:model_r2_performance_regression}
\vspace{-12pt}
\end{wraptable}
\hspace{-3pt}neighboring segments, shows a 0.13 percent point inferiority, and, unexpectedly, a 0.09 percent point lower performance as well, in comparison to the corresponding LS\textsubscript{R, no edges} model that does not incorporate information from the segments' neighborhoods. In the DV3+RN18 case, the best-performing model is the LS\textsubscript{R} GNN that is composed of a linear and GraphSAGE layer with a mean aggregation function. The model archives an average R\textsuperscript{2}-score of 78.29\%, which is slightly superior to the strong LL\textsubscript{R} baseline model.

\textbf{Setup of the Architecture Selection Process and the Hyperparameter Analysis.} A Bayesian optimization method from the SMAC3 package \cite{JMLR:v23:21-0888} was employed to identify model architectures for evaluation. This utilized the fixed 80/20 data split from the Cityscapes \cite{cordts2016cityscapes} validation set and the trained  DV3+WRN38 segmentation model. The hyperparameters of the model’s architecture that were optimized include the types of layers, the depth of the model, the width of the layers, and the learning rate. The algorithm has the option to select for each layer stage (independently) either a linear layer, that does not consider information from neighboring segments, or one of the three GNN layers: GAT, GraphSAGE with a maximum aggregation function, and GraphSAGE with a mean aggregation function, respectively. Additionally, a rectified linear unit (ReLU) activation function was incorporated between each layer of the model. The remaining hyperparameter values were constrained as follows: The maximum depth of the model architecture was fixed at four, as a GNN comprising a total of four GraphSAGE and GAT layers, respectively, incorporates four-hop neighboring features in the node embeddings (cf.\ \cite{ hamilton2017inductive, velivckovic2017graph}). We observed in our experiments that depth four is sufficient for the purposes of our investigation. Moreover, the minimum depth was set at two. The number of neurons in each layer was explored in a range between 10 and 400. Consequently, the identified model architectures through Bayesian optimization comprise a depth of up to four layers, with up to 400 neurons in each layer. Each layer is either a linear layer, a GAT layer, or a GraphSAGE layer with mean or maximum aggregation function. It should be noted, however, that in all cases the final layer is composed of a single neuron and that the subsequent activation function is always the sigmoid. Additionally, the learning rate was explored in the range of $[0.001, 0.2]$.

\textbf{Selection Process.}
From the optimization results, the best-performing model architectures were selected for each model depth value. For a given depth value, if multiple architectures obtained similar top performances, we selected two considerably different architectures among them for further experiments. Moreover, in the graph-based meta classification case, the two best-performing architectures that contain GAT layers were selected for further evaluation. Furthermore, each best-performing model for meta classification and meta regression that contains only linear layers (LLLL\textsubscript{C} and LL\textsubscript{R}) was chosen as baseline. The selected architectures are shown in \cref{tab:architectures_all}. 

\begin{figure*}[t]
\begin{center}
\begin{tabular}{cccc}
\includegraphics[width=0.38\textwidth]{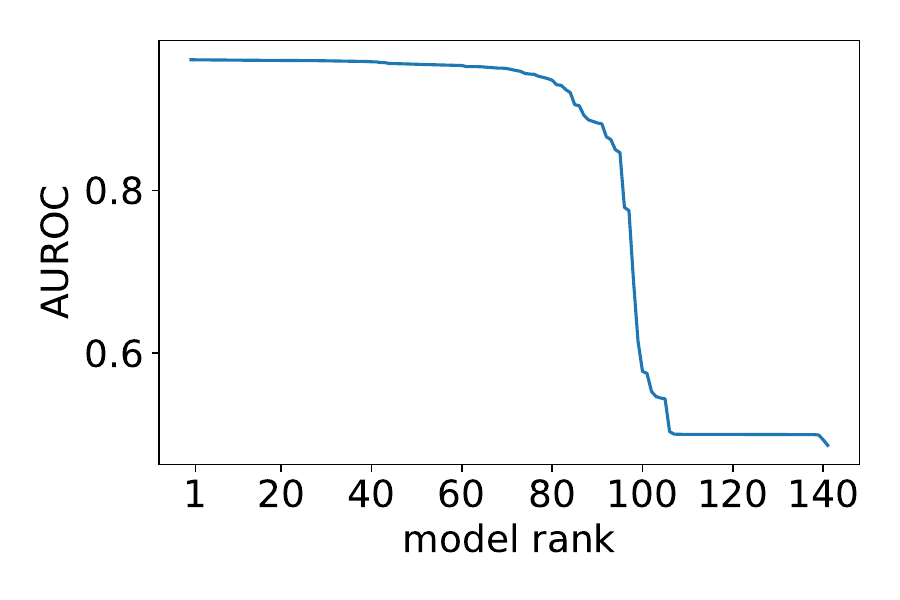}&
\includegraphics[width=0.38\textwidth]{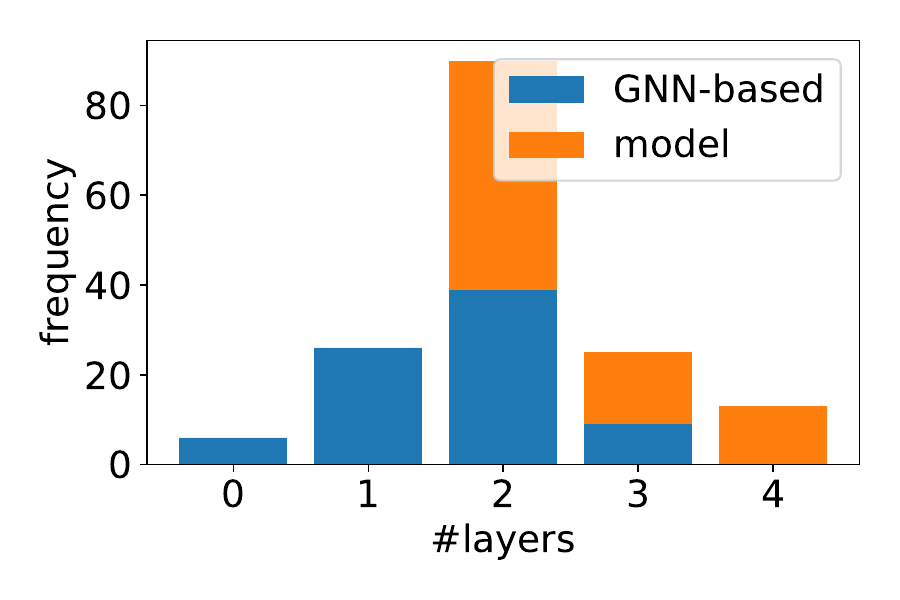}\vspace{-8pt}\\
(a)&(b)
\vspace{-2pt}\\
\includegraphics[width=0.38\textwidth]{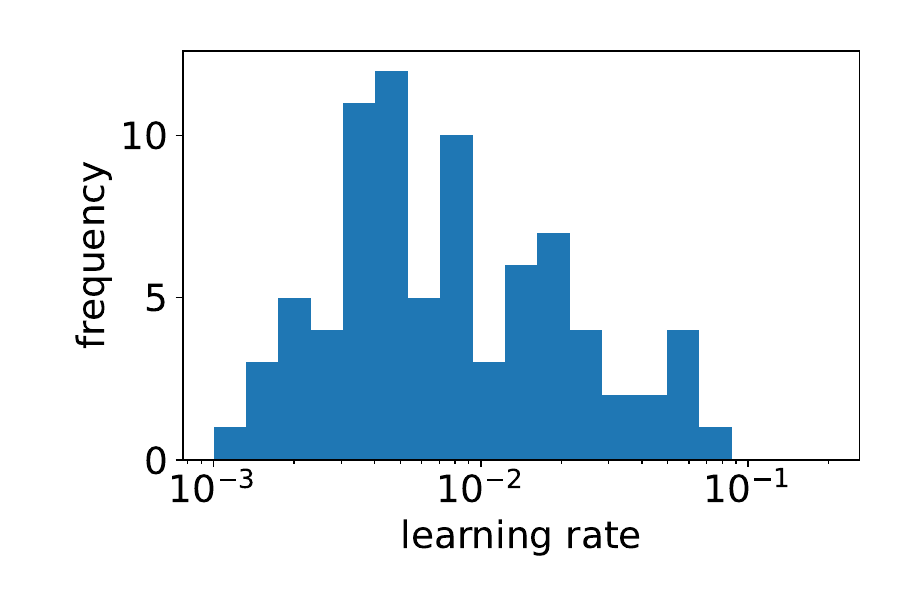}&
\includegraphics[width=0.38\textwidth]{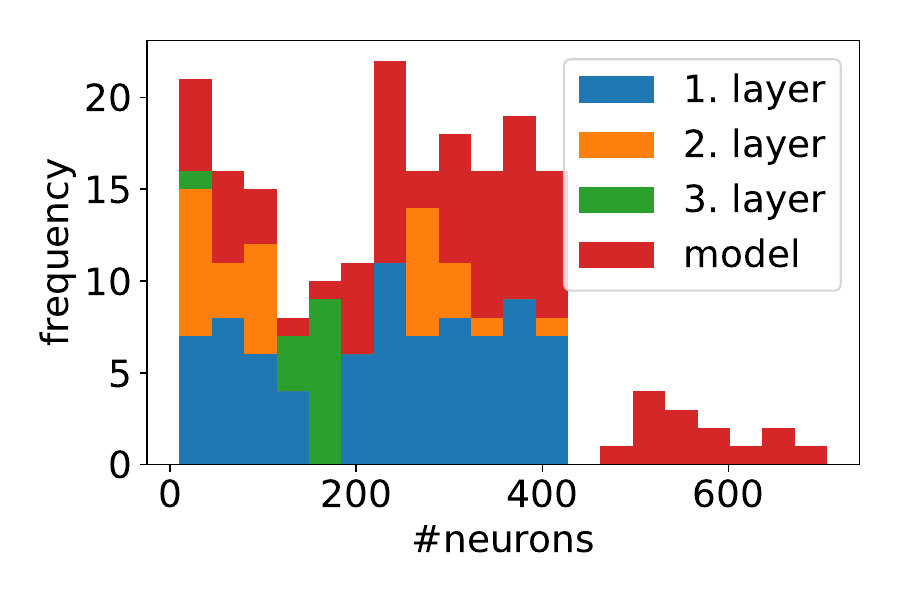}
\vspace{-8pt}\\
(c)&(d)
\vspace{-6pt}
\end{tabular}
\end{center}
\caption{Illustration of the characteristics of the Bayesian optimization results for the top 80th ranked model architectures. (a) AUROC course over the model rank (models ordered from highest to lowest AUROC). (b) histogram of the number of layers in the model (orange) and the number of GNN-based layers within the architecture (blue). (c) distribution of the learning rate of the model architectures. (d) histogram of the number of neurons in the layers (blue, orange, green), except each last layer with only one fixed neuron, and the total number of neurons in the architecture (red).}
\label{fig:architecturs_chracteristics_top_K}
\vspace{-8pt}
\end{figure*}

\textbf{Hyperparameter Analysis. }
In the course of Bayesian optimization of the graph-based meta classifiers, 142 distinct model architectures were evaluated in accordance with the methodology and definition of the hyperparameter space described before. The characteristics of the 80 best-performing models are analyzed in the following, as a notable decline in the AUROC value is observed within that region (cf.\ \cref{fig:architecturs_chracteristics_top_K}(a)).

In \cref{fig:architecturs_chracteristics_top_K}(b), a histogram of the number of layers in the model and the number of GNN-based layers within the architecture is presented. The number of layers in the model is displayed in orange, while the number of GNN-based layers within the top 80 architectures is shown in blue. Fifty-one of the top 80 ranked model architectures contain two layers. The remaining 29 are distributed approximately equally between architectures with a depth of three and four, respectively. Noteworthily, among the top 80 architectures, 65 contain one or two GNN-based layers. This observation leads to the conclusion that integrating information from one- or two-hop neighboring segments is a sufficient approach. Only nine architectures  comprise three GNN-based layers, while six architectures have no GNN-layer at all. Consequently, these latter six architectures are comprised by linear layers only. In the top 80 architectures, 66 models contain at least one GraphSAGE layer, with 40 having a mean and 26 having a maximum aggregation function. However, we found that maximum aggregation was generally less reliable than mean aggregation, frequently exhibiting a tendency to collapse in repeated training runs, resulting in an AUROC value of approximately 50\%. Therefore, such model architectures were excluded from the final selection process. Furthermore, 30 out of the 80 top-ranked architectures contain at least one GAT layer. It should be noted that the first GNN comprising at least one GAT layer is ranked 45th. Architectures with GAT layers demonstrate a comparable outcome across multiple training runs in our experiments, as evidenced by models with GraphSAGE and maximum aggregation functions (cf.\ high standard deviation in \cref{tab:model_classification_performance}).

The learning rate distribution of the top 80 architectures demonstrates a concentration around a learning rate of $3\times10^{-3}$ (cf.\ \cref{fig:architecturs_chracteristics_top_K}(c)). During the Bayesian optimization process of the top 80 architectures, no models were trained with a learning rate exceeding $8\times10^{-2}$. The distribution of the number of neurons in the first layer is approximately uniform (cf. blue bars in \cref{fig:architecturs_chracteristics_top_K}(d)), while the distributions for the second (orange bars) and third layer (green bars) are not. The former one has got two high-density intervals at approximately $[10,112]$ and $[250,325]$ and the latter one at $[125,175]$.
\section{Conclusion and Outlook}

In this work we have demonstrated that the relationships between neighboring predicted segments produced by DNNs are helpful for uncertainty estimation and prediction quality estimation. Notable performance improvements over the presented baselines were achieved. Our hyperparameter study revealed that GNNs of moderate size are sufficient to address corresponding estimation tasks. For future work it would be interesting to use GNNs not only for spatial but also temporal relationships in a unified manner.

\section*{Acknowledgment}
We would like to thank Daniel Siemssen for his invaluable contribution in transferring the base code to Python and significantly improving its efficiency. E.H.\ and M.R.\ acknowledge support by the German Federal Ministry of Education and Research within the junior research group project “UnrEAL” (grant no.\ 01IS22069). S.T.\ is funded by the VolkswagenStiftung, “DigiData” (grant no. 9C436).

\bibliography{egbib}

\end{document}